\documentclass[lettersize,journal]{IEEEtran}
\usepackage{amsmath,amsfonts}
\usepackage{algorithmic}
\usepackage{algorithm}
\usepackage{array}
\usepackage[caption=false,font=normalsize,labelfont=sf,textfont=sf]{subfig}
\usepackage{textcomp}
\usepackage{stfloats}
\usepackage{url}
\usepackage{verbatim}
\usepackage{graphicx}
\usepackage{cite}

\usepackage[utf8]{inputenc} 
\usepackage[T1]{fontenc}    
\usepackage{hyperref}       
\usepackage{url}            
\usepackage{booktabs}       
\usepackage{amsfonts}       
\usepackage{nicefrac}       
\usepackage{microtype}      
\usepackage{color}
\usepackage{bm}
\usepackage{multirow}
\usepackage{wrapfig}
\usepackage{amssymb}
\usepackage{pifont}
\newcommand{\xmark}{\ding{55}}%

\begin{document}

\title{E$^2$-AEN: End-to-End Incremental Learning with Adaptively Expandable Network}

\author{Guimei~Cao$^*$,
        Zhanzhan~Cheng$^*$,
        Yunlu~Xu$^*$,
        Duo~Li,
        Shiliang~Pu,
        Yi~Niu,
        and Fei~Wu
\IEEEcompsocitemizethanks{
\IEEEcompsocthanksitem G. Cao, Z. Cheng, Y. Xu, D. Li, S. Pu and Y. Niu are with Hikvision Research Institute, Hangzhou, 310051, China (email: [caoguimei, chengzhanzhan, xuyunlu, liduo6, pushiliang.hri, niuyi]@hikvision.com).F. Wu are with College of Computer Science and Technology, Zhejiang University, Hangzhou, 310058, China (e-mail: wufei@cs.zju.edu.cn). Z. Cheng is also with College of Computer Science and Technology, Zhejiang University, Hangzhou, China.
}

\thanks{$^*$G. Cao, Z. Cheng and Y. Xu contributed equally to this research. }
}

\maketitle

\begin{abstract}
Expandable networks have demonstrated their advantages in dealing with \emph{catastrophic forgetting} problem in incremental learning.
Considering that different tasks may need different structures,
recent methods design dynamic structures adapted to different tasks via sophisticated skills.
Their routine is to search expandable structures first and then train on the new tasks, which, however, breaks tasks into multiple training stages, leading to suboptimal or  overmuch  computational cost.
In this paper, we propose an end-to-end trainable adaptively expandable network named E$^2$-AEN, which dynamically generates lightweight structures for new tasks without any accuracy drop in previous tasks.
Specifically, the network contains a serial of powerful feature adapters for augmenting the previously learned representations to new tasks, and avoiding task interference.
These adapters are controlled via an adaptive gate-based pruning strategy which decides whether the expanded structures can be pruned, making the
network structure dynamically changeable according to the complexity of the new tasks.
Moreover, we introduce a novel sparsity-activation regularization to encourage the model to learn discriminative features with limited parameters.
E$^2$-AEN reduces cost and can be built upon any feed-forward architectures in an end-to-end manner.
Extensive experiments on both classification (\textit{i.e.}, CIFAR and VDD) and detection (\textit{i.e.}, COCO, VOC and ICCV2021 SSLAD challenge) benchmarks demonstrate the effectiveness of the proposed method, which achieves the new remarkable results.
\end{abstract}

\section{Introduction} \label{intro}
\IEEEPARstart{I}{ncremental} learning is devoted to learning new tasks/data incrementally without \emph{catastrophic forgetting}.
It has been proven useful for various online vision systems \cite{ostapenko2019learning,aljundi2019task,rajasegaran2020itaml} such as autonomous driving and robotics. 
Early methods attempt to design regularization-based strategies \cite{kirkpatrick2017overcoming,chaudhry2018riemannian,2017Continual,2018Piggyback,2018PackNet,2020Continual,HuTM0Z21,TangCZYO21} for preserving the old knowledge through important weights in the network.
Such weights will cause intransigence towards new tasks, and lead to a performance bottleneck when the number of tasks grows faster.
Then replay memory-based methods \cite{rebuffi2017icarl,lopez2017gradient,2018Lifelong,2018End,BangKY0C21,CO2L,recall2021,SimonKH21,9368260,9540230,9488623} are developed by storing a fraction of old feature representations and replaying them when retraining models for new tasks.
However, if sequential tasks are less relevant, they suffer from the stability-plasticity dilemma \cite{rajasegaran2020itaml,2013Adaptive,2021DER}.
\begin{figure*}[]
\center{\includegraphics[width=15cm] {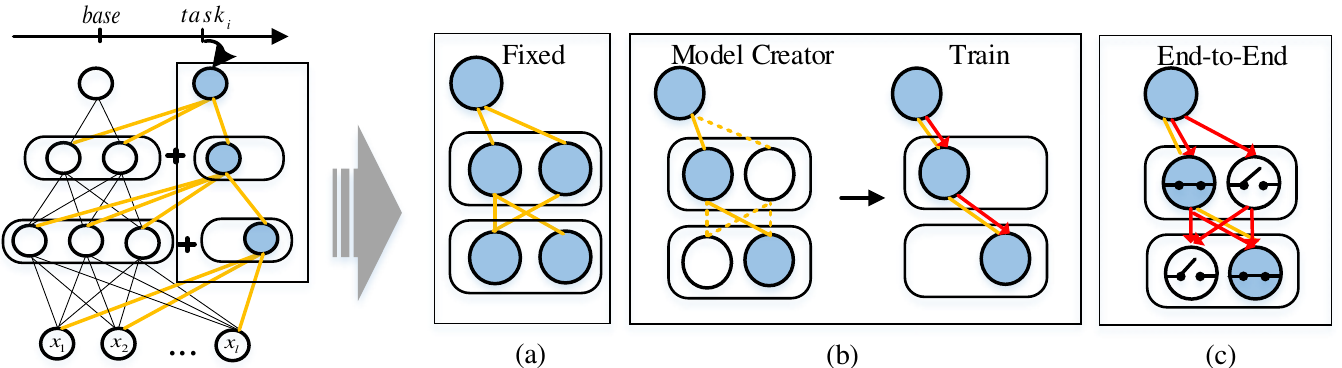}} 
\caption{\label{1} A conceptual overview of expandable networks paradigm, in which three kinds of expandable strategies can be applied.
(a) The fixed expanding strategy whose parameter scale is linearly increasing.
(b) The existing solutions can be generalized in two steps orderly: the task-specific model creation (\textit{i.e.}, using pruning or NAS-based strategy for suitable structures), and  training for the specific task using the optimized expanded networks.
(c) Our proposed end-to-end adaptively expandable network (E$^2$-AEN), which dynamically adjusts network architectures for different tasks. Yellow lines refer to connections in the forward process, and red lines refer to the backpropagation.}
\label{motivation}
\end{figure*}

{
Recently, parameter isolation \cite{delange2021continual} becomes a new trend of incremental learning since that different tasks may need different parameters or network structures, avoiding interference between tasks.
A group of mask-based methods \cite{2018PackNet,J2018Overcoming,abati2020conditional} utilize task-specific weighting strategies to freeze the important weights for given tasks.
However, the capacity of the backbone model tends to saturate if a large number of tasks are coming, limiting the rich compositionality of the incremental framework.

Another group of methods is called expandable networks\cite{rusu2016progressive,rosenfeld2018incremental,kanakis2020reparameterizing,yoon2017lifelong,li2019learn,wang2020lifelong,xu2018reinforced}.
They grow new branches for new tasks while freezing previous task parameters, allowing to expand alongside the sequential incremental tasks.
The addition of new capacity 
gives these models the flexibility to both reuse old computations and learn new ones.
Early trials \cite{rusu2016progressive,rosenfeld2018incremental,kanakis2020reparameterizing} adopt the fixed expanding structure at the cost of a linear increase of parameter memory and computational cost, as shown in Fig.  \ref{motivation} (a).
To reduce computational cost or extra parameter overhead of the fixed expandable structures, dynamic expandable strategies \cite{yoon2017lifelong,li2019learn,wang2020lifelong,xu2018reinforced, Enzo2019Neural} are introduced through fantastic design, \textit{e.g.}, pruning \cite{yoon2017lifelong}, Network Architecture Search (NAS) \cite{li2019learn,wang2020lifelong,Enzo2019Neural} or reinforcement learning \cite{xu2018reinforced}, as shown in Fig.  \ref{motivation} (b).
Unfortunately, they usually \emph{solve one problem only to find another cropping up}.
For example, \cite{yoon2017lifelong} reduces the parameter scale while relying on time-consuming retraining.
Both \cite{li2019learn} and \cite{wang2020lifelong} { apply NAS to find better structures for sequential tasks while bringing about huge searching costs.}
In addition, they all divide the learning process into multiple phases: defining the task-specific network first and then training respectively.
Then much more hyperparameters and re-training strategies are heavily dependent, resulting in overmuch computational cost yet suboptimal.
Such methods are hard to be applied to resource-constrained scenarios like mobile devices or video surveillance cameras.

{
Upon above, we believe that a good expandable network should be with three abilities.
(1) \emph{stability-plasticity}: adaptively preserving the performance of old tasks and adapting to new tasks well.
Models should be (2) parameter-efficient (\emph{lightweight}), and (3) low extra computational cost (\emph{efficiency}).
TABLE \ref{ability} shows the differences among existing expandable strategies.
}
Keep these key characteristics in mind, in this paper, we propose an \textbf{E}nd-to-\textbf{E}nd trainable \textbf{A}daptively \textbf{E}xpandable \textbf{N}etwork (E$^2$-AEN), achieving global optimization.
The proposed method can dynamically adjusts network architectures for different tasks, as illustrated in Fig. \ref{motivation} (c).
Specifically,
E$^2$-AEN has a serial of lightweight modules called Adaptively Expandable Structure (\emph{abbr.} AES) accompanied the base networks.
Each AES is corresponding to the specific task, which has two key steps:
(1) Enhancing the originally {learned} representations with a well-designed \emph{feature adapters} for adapting to new tasks. The \emph{feature adapters} are in form of residual units and can meet the \emph{stability-plasticity}.
(2) An adaptive \emph{gate-based pruning strategy} is used to decide whether the expanded structures can be pruned. It can make the network structure changeable according to the complexity of the new tasks.
For the optimization of AES, a sparsity-activation regularization is proposed to encourage AES to learn discriminative features for new tasks while reducing parameters as much as possible.
In this way, both \emph{lightweight} and \emph{efficiency} are taken into account over the entire process.
More details are shown in Section \ref{method}. 

\begin{table}
\centering
\caption{Comparisons among Different Expandable Strategies. They Are the Fixed Expanding \cite{rusu2016progressive,rosenfeld2018incremental,kanakis2020reparameterizing}, the Dynamic Expanding \cite{yoon2017lifelong,li2019learn,wang2020lifelong,xu2018reinforced,Enzo2019Neural} and the Proposed Strategy.}
\label{ability}
\setlength\tabcolsep{10pt}{
\begin{tabular}{lccc}
\toprule
Strategy & \emph{Stability-Plasticity} & \emph{Lightweight} & \emph{Efficiency} \\ \midrule
(a) Fixed &  \checkmark       &\xmark        &\xmark    \\ \midrule
(b) Dynamic  & \checkmark   &\checkmark &\xmark  \\ \midrule
(c) E$^2$-AEN          & \checkmark     & \checkmark & \checkmark            \\ \bottomrule
\end{tabular}
}
\end{table}

Our main contributions are summarized as follows:
(1) We propose a trainable adaptively expandable network, which preserves the existing task performance through freezing isolated parameters, and achieves promising performance on new tasks via the well-designed adapters.
 {In addition, it is lightweight and easily optimized in an end-to-end trainable manner.}
(2) We implement it with a serial of {Adaptively Expandable Structures (AES)} which are composed of powerful \emph{feature adapters} and an adaptive \emph{gate-based} \emph{pruning strategy}.
Moreover, a novel sparsity-activation constraint is applied for balancing the performance and parameter numbers. 
(3) Extensive experiments on both classification (CIFAR and VDD) and detection (VOC, COCO and ICCV2021 SSLAD competition) datasets demonstrate the superiority of our proposed approach over the state-of-the-art.

\section{Related Work} \label{related}

\subsection{Incremental Learning}
Incremental Learning (IL) aims at designing new learning schemes to learn new knowledge without forgetting old tasks, in which \emph{catastrophic forgetting} is the key researching problem.
To overcome
\emph{catastrophic forgetting} problem, many approaches have been proposed, roughly divided into three categories: regularization-based \cite{kirkpatrick2017overcoming,chaudhry2018riemannian,2017Continual,2018Piggyback,2018PackNet,2020Continual,HuTM0Z21,TangCZYO21},
replay memory-based \cite{rebuffi2017icarl,lopez2017gradient,2018Lifelong,2018End,BangKY0C21,CO2L,recall2021,SimonKH21,9368260,9540230,9488623} and parameter isolation-based \cite{rosenfeld2018incremental,kanakis2020reparameterizing,abati2020conditional,rusu2016progressive,yoon2017lifelong,li2019learn,wang2020lifelong,xu2018reinforced} methods. Our approach belongs to the \textit{parameter isolation-based} paradigm,
which dedicates different model parameters to each task, to prevent any possible forgetting.

In this family,
one intuitive way is to select different sub-networks from a single fixed network for respective tasks.
\cite{PathNet} selected a path for the current new task by a tournament selection genetic algorithm.
\cite{2018PackNet} identified the important weights for prior tasks by pruning, and then retrained with the less relevant subset of weights. This results in a lower number of parameters being free and performance dropping quickly on longer sequences.
\cite{J2018Overcoming} used incorporating task-specific embedding for attention masking to imply the importance of weights for tasks. 
Later, \cite{abati2020conditional} developed the task-dedicated gating modules that select which filters to apply conditioned on the input feature,
which practically generated a channel-wise binary mask in every layer for each task.

As the number of tasks increases, a fixed network with a limited capacity {will struggle to overcome the stability-plasticity dilemma} \cite{rajasegaran2020itaml,2013Adaptive,2021DER}, similar to the regularization-based and replay memory-based methods.
\cite{rusu2016progressive} proposed the progressive network to assign additional fixed model resources to each task. When a new task comes, the parameters of the old model will be frozen and the new model modules will be trained.
\cite{rosenfeld2018incremental} proposed a deep adaptation network (DAN) that constrained newly learned filters to be linear combinations of existing frozen ones, which can preserve the performance on the original domain.  \cite{2017LearningAdapters} introduced additional parametric convolutional layers to modify a standard residual network adapting to new tasks.
\cite{kanakis2020reparameterizing} reparameterized the convolutional network into a non-trainable shared filter bank and task-specific parts.
However, the number of model parameters in the aforementioned methods tends to increase rapidly as the number of tasks increases.
Recently, some dynamic expandable methods \cite{yoon2017lifelong,xu2018reinforced,li2019learn,wang2020lifelong} took both performance and computation cost (\emph{e.g}, the growth of model size) into consideration. 
\cite{yoon2017lifelong} proposed a dynamically expandable network (DEN), that can automatically decide its network capacity as it trains on a sequence of tasks.
\cite{xu2018reinforced} assigned different new modules for different tasks by a controller trained by reinforcement learning.
Both \cite{li2019learn} and \cite{wang2020lifelong} used NAS to find better structures for the sequential tasks, followed by parameter retraining for the final results.
{
Unfortunately, all searching-and-training strategies are time-consuming and computationally expensive.
Then two adaptive methods \cite{2021DER,LiuSS21} are recently proposed to alleviate the stability-plasticity dilemma relying on old data.
}
Unlike the above methods, our method can generate lightweight structures adaptively, achieving a global optimization from scratch without accessing the old data. 


\begin{table*}
\centering
\caption{Comparison between Incremental Learning and Several Related Topics. ``$\mathcal{P}^{Src}_{x,y}$ vs. $\mathcal{P}^{Tar}_{x,y}$'' Refers to the Source/Target Data Distribution. ``$\mathcal{Y}^{Src}$ vs. $\mathcal{Y}^{Tar}$'' Means the Source/Target Label Space. $\mathcal{S}_i$ Is the i-th Subset.}
\label{topics comparison}
\setlength\tabcolsep{10pt}{
\begin{tabular}{lccccc}
\toprule
Learning paradigm    & Training data & Test data & $\mathcal{P}^{Src}_{x,y}$ vs. $\mathcal{P}^{Tar}_{x,y}$ & $\mathcal{Y}^{Src}$ vs. $\mathcal{Y}^{Tar}$ & Condition \\ \midrule
Multi-task learning (ML)  & $\mathcal{S}_1 \ldots \mathcal{S}_n$     &$\mathcal{S}_1 \ldots \mathcal{S}_n$        & $=$ & $\neq$     & Each subset is available           \\ \midrule
Transfer learning (TL)    & $\mathcal{S}_{src}, \mathcal{S}_{tar}$     & $\mathcal{S}_{tar}$       &$\neq$  & $\neq$     &  -         \\ \midrule
Domain adaptation (DA)    & $\mathcal{S}_{src}, \mathcal{S}_{tar}$   &$\mathcal{S}_{tar}$        & $\neq$ & $=$     &  -        \\ \midrule
Incremental learning (IL)  & $\mathcal{S}_1 \ldots \mathcal{S}_n$   &$\mathcal{S}_1 \ldots \mathcal{S}_n$        & $\neq$  &  $\neq$      & $\mathcal{S}_i$ arrives sequentially          \\ \bottomrule
\end{tabular}
}
\end{table*}

\begin{figure*}[]
\center{\includegraphics[width=17cm] {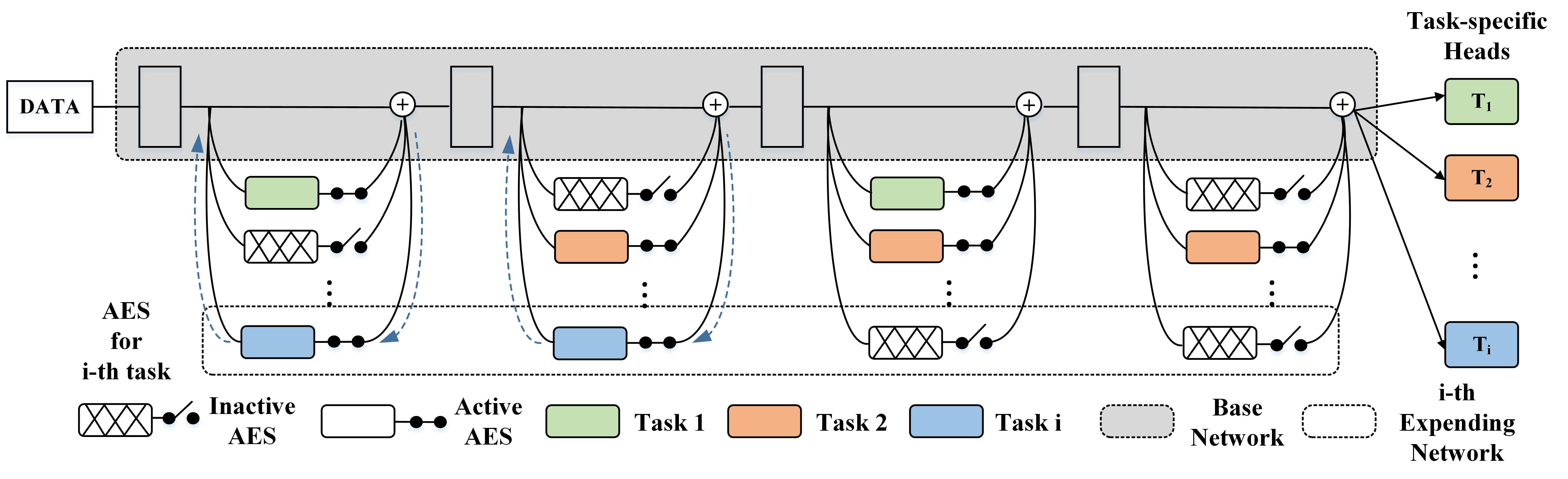}}
\caption{\label{1} Illustration of the proposed E$^2$-AEN. Its essential AES makes the existing convolutional layers adapt to new tasks, and back-propagation is only conducted on the task-specific units of current tasks, while preserving the parameters of previous tasks.
}
\label{framework}
\end{figure*}

\subsection{Related Topics}
There are several research topics \cite{LLOQ21} closely related to incremental learning.
They are multi-task learning, transfer learning and domain adaptation.
We summarize them in TABLE \ref{topics comparison}, and give descriptions as follows.

\textbf{Multi-Task Learning (MTL)}\cite{2020Multi} focuses on learning multiple tasks simultaneously, and tends to solve this problem from both architecture design \cite{adashare, MTAN, NDDR, VandenhendeGGB20, LuKZCJF17} and loss functions \cite{uncertainty, gradnorm, GuoHHYF18, SenerK18} perspectives.
Though it has high-relation with IL on the data setting (training and test), they differ mainly in that: Firstly, the goal of IL is to sequentially learn models for a sequence of tasks independently. It means only the current task ($\mathcal{S}_{i}$) is available for training. While MTL relies on the entire dataset and jointly training.
Secondly, IL methods make effort to preserve knowledge learned on previous tasks when coping with new tasks, while MTL methods struggle to decide what parameters to share across all tasks.

\textbf{Transfer Learning (TL)} \cite{PanY10, TanSKZYL18, ZhuangQDXZZXH21} aims to transfer the knowledge from one (or multiple) source task/domain to a different but related task/domain.
Pretraining-finetuning is a common transfer learning strategy.
In this setting, the user should train the model on a large scale dataset firstly such as ImageNet \cite{2014ImageNet}, and then finetune the model on the downstream tasks.
TL is only concerned with the ability for transferring knowledge to the target task.
However, IL should have the ability to accommodate new knowledge continually, and also preserve the learned knowledge.

\textbf{Domain Adaptation (DA)}  \cite{PatelGLC15, WangD18} is also a popular topic recently. DA aims to maximize performance on the target domain using a source domain.
The difference between DA and TL is on the label space.
The differences between DA and IL are, (1) DA is available of the existing source domain and target domain while IL is not available of previous tasks/domains, and (2) DA cares merely about the performance on target domain while IL concerns the performance on both current and previous tasks/domains.

In this paper, we focus on the IL task, and should make comparison with the IL-based solutions.

\section{Method} \label{method}
\subsection{Problem Formulation}
We consider a task-incremental problem under the incremental learning scenario, where a sequence of tasks $\{task_1, \ldots, task_i, \ldots, task_N\}$ are given to a parametric model.
For each task, its training data is formulated as $D_i=\{{{\bm x}_j, y_j}\}^{N_i}_{j=1}$ where ${\bm x}_j$ is the original input data, $y_j$ is the corresponding label, and $N_i$ refers to the example scale of the $i$-th task.
To learn the $i$-th task incrementally, we aim to optimize a deep function $y_i=\Phi\left ({\bm x}_i;  \Theta^s,  \Theta_{i}^t \right )$ where $ \Theta^s$ and $ \Theta^t$ are the task-shared and task-specific parameters, respectively.
Then its objective function is formulated as:
\begin{equation}
\mathcal{L} =  \sum_{i=1}^N \mathcal{L}_i\left( { \Theta_i^t}; { \Theta^s}, D_i\right),
\end{equation}
where $\mathcal{L}_i$ identifies the  loss function of $i$-th task.
Conventionally, the task-id of the given task is predefined and available.

\subsection{Adaptively Expanding Mechanism}
{
Our proposed \textbf{E}nd-to-\textbf{E}nd trainable \textbf{A}daptively \textbf{E}xpandable \textbf{N}etwork (E$^2$-AEN) can supersede the stage-wise optimization attributing to the essential adaptively expanding mechanism. It is realized through a serial of lightweight modules called Adaptively Expandable Structure (AES) accompanied the base networks,
corresponding to each specific task as shown in Fig. \ref{framework}.}
AES has two possible states, active and inactive.
When the state is active, the \emph{feature adapter} is equipped with the frozen backbone for adapting to the new task.
Otherwise, the \emph{feature adapter} is skipped thus saving some unnecessary computation and parameters, which is adaptive and efficient.

Given the $j$-th convolutional block in the base network denoted as $H_{j}$,
the $j$-th convolutional block in the $i$-th incremental task $F_{i,j}$ through AES forward pass is
\begin{equation} \label{acm}
F_{i,j}= H_{j}+\mathcal{A}_{i,j}(H_{j}).
\end{equation}
Here $\mathcal{A}(\cdot)$ is a differentiable module in our proposed AES.

The proposed AES has two key components: the designed \emph{feature adapter} and the adaptive \emph{gate-based pruning strategy}, as specified in Fig. \ref{aes}.
The powerful \emph{feature adapter}, whose operation is denoted as $\mathcal{D}(\cdot)$, is responsible for augmenting the previously learned representations to new concepts, and avoiding task interference effectively. The backbone parameters are frozen so that there is no forgetting phenomenon on prior tasks.
The adaptive \emph{gate-based pruning strategy} is dedicated to reducing the redundant parameters brought by the \emph{feature adapter}, making the framework more efficient, whose operation is denoted as $\mathcal{G}(\cdot)$.
Profit from \emph{gate-based pruning strategy}, AES is capable of determining whether the execution of the \emph{feature adapter} in a certain convolutional layer is required or it can be skipped.
Then (\ref{acm}) turns to
\begin{equation} \label{acm2}
F_{i,j}=H_{j}+\mathcal{G}_{i,j}(\mathcal{D}_{i,j}(H_{j}))\cdot \mathcal{D}_{i,j}(H_{j}),
\end{equation}
where $\mathcal{G}_{i,j}(\cdot) \in \{0,1\}$ and $\cdot$ means multiplication.

\subsubsection{\bf Feature Adapter}
In general, AES works as the task-specific part in the whole incremental framework, where \emph{feature adapter} plays an essential role in acquiring new concepts.
There are two na{\"i}ve implements for these adapters.
\cite{rosenfeld2018incremental} proposed a reparameterization to constrain the newly learned filters to be a linear combination of existing $1\times 1$ convolutional filters, denoted as \emph{plain}$_{1\times 1}$.
Instead of changing the filter coefficients directly, \cite{2017Learning} introduced the residual architecture for a better feature presentation, denoted as \emph{residual}$_{1\times 1}$.
Considering the efficiency, neither of the above injected nonlinear modules or more complex structures into these feature adapters, na{\"i}ve adapters lead to the limited capability of adapted feature representation.
{
We believe that the ideal \emph{feature adapter} should be capable of learning powerful representation while keeping it lightweight.
Hence, we design an effective \emph{feature adapter} by introducing the diverse size of filters and multiple nonlinear units. 
In this way, the model can learn diverse and discriminative features for new concepts without bringing in much additional cost}.
Specifically, at $j$-th convolutional block of $i$-th task, \emph{feature adapter} is denoted as $\mathcal{D}_{i,j}(\cdot)$, containing 3 convolutional layers with filter sizes of $1 \times 1$, $3 \times 3$  and $1 \times 1$, respectively.
Formally, the output of the \emph{feature adapter} is denoted as:

 \begin{equation}
 \begin{aligned}
 E_{i,j}&=H_j + \mathcal{D}_{i,j}(H_j) \\
      &=H_j +  \sigma \left ( {\Theta}^{t,3}_{i,j} \otimes \sigma \left ( {\Theta}^{t,2}_{i,j} \otimes \sigma \left ( {\Theta}^{t,1}_{i,j}\otimes H_j \right ) \right ) \right ),
 \end{aligned}
 \end{equation}
where the channel size of $H_j$ is denoted as $c$, and ${\Theta}^{t,1}_{i,j} \in \mathbb{R}^{c\times \frac{c}{3}}$, ${\Theta}^{t,2}_{i,j} \in \mathbb{R}^{\frac{c}{3} \times \frac{c}{6}}$, ${\Theta}^{t,3}_{i,j} \in \mathbb{R}^{\frac{c}{6}\times c}$. $\sigma(\cdot)$ is the ReLU operator. $\otimes$ means convolutional operation here.
Both the kernel size of ${\Theta}^{t,1}_{i,j}$ and ${\Theta}^{t,3}_{i,j}$ are $1 \times 1$, the size of ${\Theta}^{t,2}_{i,j}$ is $3 \times 3$.
Thus, the added extra parameters are calculated as $c^2/3+c^2/2+c^2/6=c^2$ theoretically, which is the same as \cite{rosenfeld2018incremental,2017Learning}.
Notice that Batch Normalization is applied after each convolution for the domain adaptation.

\begin{figure}
\center{\includegraphics[width=8.5cm] {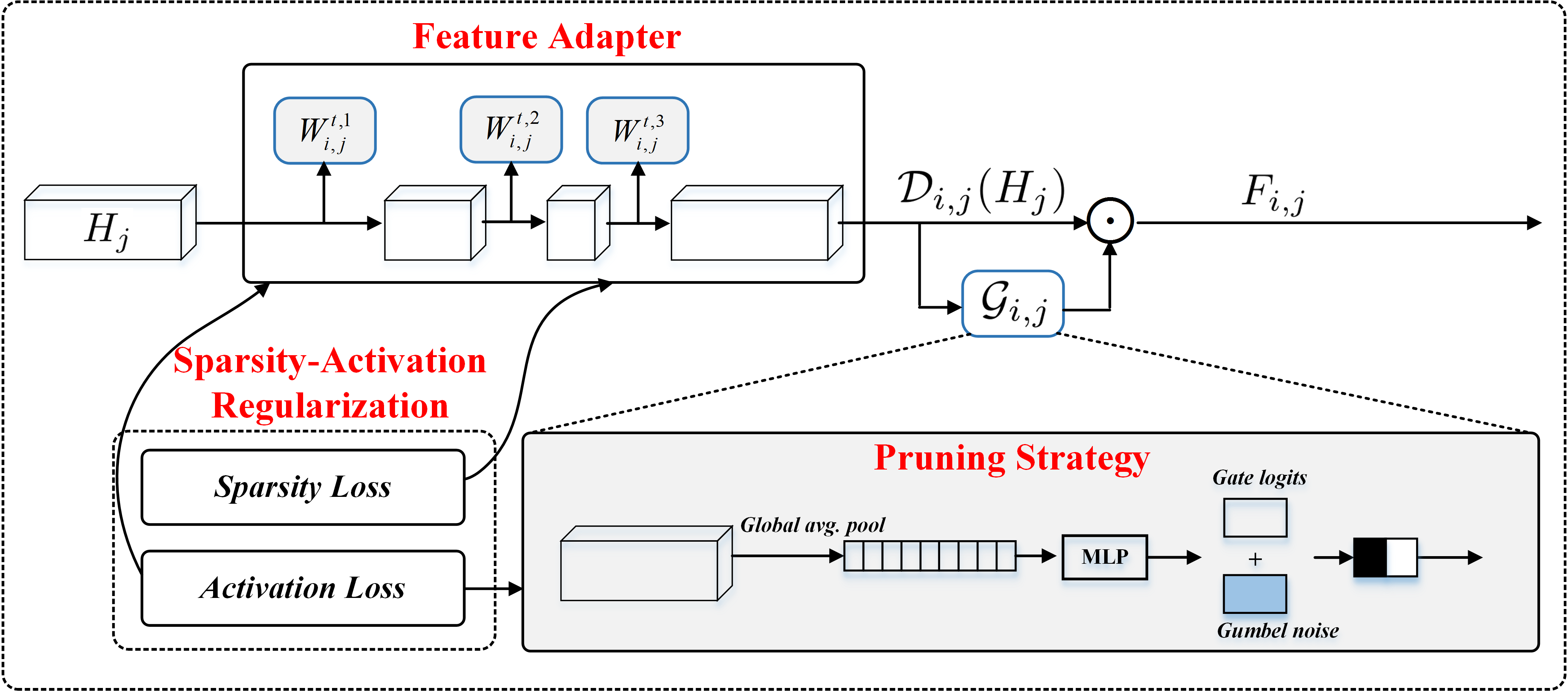}}

\caption{\label{1} Illustration of Adaptively Expanding Structure. It has two essential components, with
\emph{feature adapter} responsible for adapting to new specific tasks and \emph{gate-based pruning strategy} determining whether the execution of corresponding \emph{feature adapter} is necessary. Sparsity-activation regularization is to assist the joint learning of the two parts. Notably, the skipped \emph{feature adapter} in the training phase can be pruned exactly in the inference without any performance drop.
}
\label{aes}
\end{figure}
\subsubsection{\bf Adaptive Gate-Based Pruning Strategy}
In the feed-forward process, \emph{gate-based pruning strategy} learns to understand the features of
\emph{feature adapters}
through a global average pooling \cite{2017Squeeze}, to capture rich information in convolutional features while remaining computationally efficient.
Then two fully-connected layers connected with a ReLU activation function are followed, obtaining $ \upsilon $.
In AES, we learn the weights of \emph{gate-based pruning strategy} and \emph{feature adapters}, denoted as $\mathcal G$ and $\mathcal D$ respectively, jointly through the standard back-propagation from our well-designed loss functions. However, considering the non-differentiable binary gating operations, we adopt a Gumbel-Softmax estimator \cite{2016Categorical} to compute the biased estimate of the gradients and enable direct optimization of the discrete operation using back-propagation.
Specifically, the Gumbel-Max trick \cite{1954Statistical} provides the ability of drawing samples $z_{i,j}$ from a binary energy distribution $ \upsilon_{i,j} $:

\begin{equation}
z_{i,j}= \underset{k \in \{0,1\}}{\arg\max}\left [ \log{\upsilon_{i,j}\left( k\right )}+ g_{i,j}\left( k\right ) \right ] ,
\end{equation}
where $k$ refers to a state of the gate that 0 means inactive and 1 means active. $\upsilon_{i,j}(k)$ means the probability of $k$, which is an on-off variable with probabilities of $[1-\alpha_{i,j}, \alpha_{i,j}]$, and $\alpha_{i,j}$ represents the probability that the $j$-th block is selected to execute in the $i$-th task.
$g_{i,j}$ is a standard Gumbel distribution with $u_{i,j}$ sampled from Gumbel $\left ( 0, 1 \right )$, where $u_{i,j} \sim \rm uniform \left ( 0,1 \right)$, then $g_{i,j}$ is computed as $-\log\left(-\log\left(u_{i,j}\right)\right)$.
The Gumbel-Softmax trick relaxes one-hot $(z_{i,j}) \in \left \{ 0,1 \right \}$ (the one-hot encoding of $z_{i,j}$) as a continuous, differentiable approximation:
\begin{equation}
y_{i,j}\left(k\right)=\frac{\exp\left ( \left ( \log{\upsilon_{i,j}\left(k\right)}+g_{i,j}\left(k\right) \right )/\tau  \right )}{\underset{l\in \{0,1\}}{\sum}\exp\left ( \left ( \log{\upsilon_{i,j}\left(l\right)}+g_{i,j}\left(l\right) \right ) /\tau \right )},
\end{equation}
where $\tau$ means {a temperature (default by 1.0) of softmax}.
Since the Gumbel-Softmax distribution is smooth for $\tau > 0$ \cite{2016Categorical}, and therefore a well-defined gradient $\frac{\partial y_{i,j}}{\partial \upsilon_{i,j}}$ can be computed.
Thanks to the Gumbel-Softmax estimator, it is possible to make the hard 0-1 threshold in the forward pass while keeping it differentiable in the backward pass.
\subsubsection{\bf Sparsity-Activation Regularization}
Last, but definitely not least is the design of sparsity-activation regularization, which makes an end-to-end optimization of our method possible.
Task-specific losses only optimize for accuracy without taking efficiency into account. However, we prefer to form a compact sub-network for each task, pruning redundant expandable \emph{feature adapters} as much as possible. To achieve this goal, we firstly propose a sparsity regularization $\mathcal {L}_{\rm sparse}$ {to reduce the  feature redundancy of each feature adapter.} 
{
Nevertheless, the effect of sparsity degree on performance is changeable among various tasks, and thus how to prune those feature adapters is unclear.
A general usage in model compression is to manually set a fixed pruning threshold according to experience or observations via several prune-retrain circles. We here aim to automatically prune the \textit{feature adapters}. Thus we design an activation regularization $\mathcal {L}_{\rm active}$ to set up an inherent relation between \emph{feature adapters} and \emph{adaptive pruning strategy}, that the pruning strategy tends to prune those \textit{feature adapters} with lower sparsity, and adaptively learns a balance between the sparsity and performance.}

\textbf{Sparsity Constraint}. Firstly, the group lasso regularization \cite{2013A} is used in each \emph{feature adapter}.
Unlike the sparsity regularization used in most network pruning methods \cite{2017LearningEfficient, 2018Attention}, we consider each \emph{feature adapter} as a single group. It means the parameters in a \emph{feature adapter} will play a similar role in AES.
Its objective function is described as follows:

\begin{equation} \label{lsparse}
\mathcal{L}_{\rm sparse}=\sum_{j=1}^L \sqrt{\sum_{k=1}^{|\Omega_j|} (\theta_{j,k})^2},
\end{equation}
where $L$ refers to the number of \emph{feature adapters} in $i$-th task, 
{and we discard the symbol $i$ for clear expression as only the feature adapters belonging to $i$-th task will be optimized.}
{$\Omega_j$ means the total weights in $j$-th layer, and $\theta_{j,k}$ is the $k$-th weight in $\Omega_j$.} Concretely, $\Omega_j=\{\Theta_j^{t,1}, \Theta_j^{t,2}, \Theta_j^{t,3}\}$,
 and $|\Omega_j|$ means the number of weights in $\Omega_j$.

\textbf{Activation Constraint}.
The \emph{gate-based pruning strategy} determines whether each \emph{feature adapter} is necessary for a certain layer according to the complexity of the new
adapted features, \textit{i.e.}, an adapted feature will be simple when the sparsity of its corresponding \emph{feature adapter} is very small, thus the pruning of this \emph{feature adapter} will not bring about much performance drop indeed.
In other words, the gate $\mathcal{G}_{i,j}$ in (\ref{acm2}) should be inactivated when the \emph{feature adapter} is very sparse, vice verse.
The sparsity of $j$-th layer is defined as the ratio of the active weights.
\begin{equation}
\mathcal{R}_j^s=\mathbb{P}[|\theta_{j,k}|\geq\sigma], k \in 1,2,\ldots, |\Omega_{j}|,
\end{equation}
where $\mathbb{P}[\cdot]$
indicates the probability of those weights in $\Omega_j$ whose absolute values are larger than a threshold.
Moreover, the activation is defined as the activating ratio ($\mathcal{R}_j^a$) of samples in a mini-batch data,
\begin{equation}
\label{active}
\mathcal{R}_j^a=\frac{\sum_{k=1}^{|B|} \mathcal{G}_{i,j}^k(D_{i,j}(H_j^k))}{|B|},
\end{equation}
where $|B|$ is the number of samples in a batch, and $k$ is the sample index.
The activation regularization is represented as
\begin{equation}
\mathcal{L}_{\rm active}=\frac{1}{L}\sum_{j=1}^L {\left \| \mathcal{R}_j^s-\mathcal{R}_j^a \right \|}_2^2.
\end{equation}
Therefore, for $i$-th task, the overall loss function is
\begin{equation}
\label{loss}
\mathcal{L}_i=\mathcal{L}_{\rm ce}+\lambda_s \mathcal{L}_{\rm sparse}+ \lambda_a \mathcal{L}_{\rm active},
\end{equation}
where $\mathcal{L}_{\rm ce}$ means the cross-entropy loss in classification tasks, ${\lambda}_s$ and ${\lambda}_a$ are tunable parameters.

\subsection{Training and Inference}
Given an incremental framework with $L$ \emph{feature adapters}, different \emph{feature adapters} should adaptively have different degrees of sparsity regularization, \textit{i.e.}, ${\bm{\lambda}}^{\prime}_s=[{\lambda}^{\prime}_1, ..., {\lambda}^{\prime}_j, ..., {\lambda}^{\prime}_L]$.
Inspired by regularization-based incremental learning methods \cite{kirkpatrick2017overcoming, 2017ContinualSI} that more large gradients often refer to more important parameters,
for the $j$-th block, its sparsity degree should be adaptively tuned as ${\lambda}^\prime_j=\sum_{k=1}^{|\Omega_j|} { |\frac{\partial y_i}{\partial {\theta}_{j,k}} |}$.
Then ${\lambda}_j^\prime$ is further normalized by ${\lambda}^\prime_j=1-\frac{{\lambda}^\prime_j-\min({\bm{\lambda}}^{\prime}_s)}{\max({\bm{\lambda}}^{\prime}_s)-\min({\bm{\lambda}}^{\prime}_s)}$.
Finally, we rewrite (\ref{loss}) by replacing the scalar parameter $\lambda_s$ with ${\bm{\lambda}}_s^\prime\in \mathbb{R}^{L}$ which provides different values for each layer.

At the end of the optimization of $i$-th task, we compute the activating ratio of the gate in $j$-th layer on the total validation dataset $T_i^{val}$, which is formulated as:
\begin{equation}
R_j^a=\frac{\sum_{k=1}^{|T_i^{val}|}{G_{i,j}^k(D_{i,j}(H_j^k))}}{|T_i^{val}|}.
\end{equation}
Then we prune those feature adapters and their corresponding gates with $R_j^a$ are zero for no performance drop concern. In the
inference, the model is the final model after pruning.

\section{Experiments}

\begin{table*}[t]
  \fontsize{8}{8}\selectfont
  \centering
  \caption{Accuracy of Different Modules on CIFAR-10. FA Means Feature Adapters That Are Used and PS Means Gate-Based Pruning Strategy Is Applied in E$^2$-AEN. "$*$" Is a Symbol Means the Module Is Equipped with Its Corresponding Regularization Loss Such as $\mathcal{L}_{\rm sparse}$ or $\mathcal{L}_{\rm active}$. $r_s$ Is the Ratio of Feature Adapters with Sparsity Larger than $10\%$ Over All Feature Adapters, and $n_a$ Is the Active Numbers of AES. Both $r_s$ and $n_a$ Are Computed on T2, and the Results on Other Tasks Are Similar.}
  \label{task-specific-modules}
  \setlength\tabcolsep{10pt}{
  \begin{tabular}{lcccc|ccccc|cc}
    \toprule
    Models                                  &FA             &$\mathcal{L}_{\rm sparse}$    &PS  & $\mathcal{L}_{\rm active}$   & T1(\%)    & T2(\%)   & T3(\%)    &T4(\%)    & T5(\%)  &$r_s$(\%) & $n_a$  \\
    \midrule
    Frozen Feature                           &                 &            &              &           & 84.9   & 71.6  &75.6  &81.4 & 82.6 & -   & -         \\
    FA                    &\checkmark       &            &              &             & 99.4   & 95.9  &97.2  &{\bf 99.4} &98.7 & 82.4   &17  \\ 
    FA$^*$                     &\checkmark       &\checkmark            &       &          & 99.3   &95.8&{\bf 97.4}&99.3& 98.8  &11.8  &17    \\ 
    {FA+PS}                      &\checkmark & &\checkmark & &99.5&95.9&97.1&99.4&98.8&70.6 &17 \\

    FA$^*$+PS                   &\checkmark      &\checkmark&\checkmark                     &    & 99.2  &95.7    &97.3&99.3& 98.7    &11.8 &13   \\

    {FA+PS$^*$}                  &\checkmark &  &\checkmark & \checkmark &99.3&95.8&97.2&99.3&98.8&70.6 &12  \\

    FA$^*$+PS$^*$                    &\checkmark   &\checkmark    &\checkmark    &\checkmark     & {\bf99.4} & {\bf 96.3}  &97.3  &99.2 &{\bf 98.8} &{\bf 11.8} &{\bf 2}                 \\
    \bottomrule
  \end{tabular}}
\end{table*}

\subsection{Datasets and Network Settings}
\subsubsection{\bf Datasets}
Our method is first evaluated on Split CIFAR-10 \cite{2009Learning} and Visual Decathlon Dataset (VDD).
The Split CIFAR-10 has 5 subsets of consecutive classes, which refers to 5 binary classification tasks that are observed sequentially.
VDD consists of 10 image classification tasks, \textit{i.e.}, ImageNet \cite{2014ImageNet}, CIFAR100 \cite{2009Learning}, Aircraft \cite{2013Fine}, DPed \cite{Munder2006An}, Textures \cite{2013Describing}, GTSRB \cite{2012Man}, Omniglot \cite{2015Human}, SVHN \cite{2011Reading}, UCF101 \cite{2012UCF101} and VGG-Flowers \cite{2008Automated}.
The short side of each image is resized as 72 pixels.
Furthermore, we also evaluate our method on COCO \cite{2014Microsoft} as well as VOC \cite{2010The} as the extension.

\subsubsection{\bf Network Settings}
Following \cite{abati2020conditional}, we employ ResNet-18 \cite{2016Deep} as the backbone for Split CIFAR-10.
For VDD, we apply E$^2$-AEN built on Wide Residual Network with a depth of 28, a widening factor of 4, and a stride of 2 in the first convolutional layer of each block.
Each residual unit in a block contains a $3\times3$ convolutional layer, as well as the BN and ReLU operations.
The channel sizes of the three blocks are 64, 128, 256, respectively.
For object detection, we built upon Faster RCNN \cite{2017Faster} with SE-Resnet50 \cite{2017Squeeze}.
${\lambda}_s$ is set through the adaptive parameter tuning. ${\lambda}_a$ is selected from $\{0.005, 0.001, 0.0005, 0.0001\}$, which yields the best results.
Threshold $\sigma$ is not sensitive to the performance selected from 0.001 to 0.05, which is 0.01 by default.

\subsection{Ablation Study}

\subsubsection{\bf Effects of AES}
We explore the effects of \emph{feature adapters} and \emph{gate-based pruning strategy} in AES separately on CIFAR-10, as shown in TABLE \ref{task-specific-modules}.
Here, we consider \emph{Frozen Feature} as our baseline, \emph{i.e.}, freezing the parameters of both convolutional and BN layers of the ResNet-18 backbone.
Experiments demonstrate the effectiveness of \emph{feature adapters}, \emph{gate-based pruning strategy}, $\mathcal{L}_{\rm sparse}$ and $\mathcal{L}_{\rm active}$.
With \emph{feature adapters}, our approach FA (w/o $\mathcal{L}_{\rm sparse}$) significantly improves the baseline in 5 tasks.
$\mathcal{L}_{\rm sparse}$ largely reduces the number of \emph{feature adapters} of which the sparsity is larger than 10\%.
Moreover, there are 13 active AES without $\mathcal{L}_{\rm active}$, while this number is reduced to 2 when $\mathcal{L}_{\rm active}$ is applied.
Hence, E$^2$-AEN ensures the performance and efficiency of the incremental learning system.
\begin{table*}[htp]
  \centering
  \caption{Effects of Feature Adapters on Split CIFAR-10. \#P Means the Parameters($\times 10^6$) of the Dynamic Network.}
  \label{adapter-abla}
  \setlength\tabcolsep{12pt}{
  \begin{tabular}{lcccccccc}
    \toprule
    Feature Adapter  & T1  & T2  & T3  & T4  & T5  & AVG  & \#P  \\
    \midrule
    plain$_{1\times1}$\cite{rosenfeld2018incremental}  & 99.1 & 91.8   & 95.4    & 98.4  & 97.7  & 96.5      & 18.6         \\
    residual$_{1\times1}$\cite{2017Learning}       & 99.3  & 94.9  & 96.2    & 98.6  & 98.3  & 97.5   & 18.6      \\
    adapter in E$^2$-AEN            & \textbf{99.4}  & \textbf{95.9}  & \textbf{96.9}  & \textbf{99.4}  & \textbf{98.7}  & \textbf{98.1} & 18.6        \\
    \bottomrule
  \end{tabular}
  }
\end{table*}

\begin{table*}[]
  \centering
  \caption{The Stability of Gate-Based Pruning Strategy on VDD. For the  Given Task, Gate-Based Pruning Strategy (Denoted as \emph{PS}) Always Adaptively Activates the Suitable Blocks, and Obtain the Robust Performance, Which Obviously Outperforms the Random Settings. Acitive\_Pos Means the Active AES. }
  \label{active-gate-pos-analysis}
  \setlength\tabcolsep{6pt}{
  \begin{tabular}{c|c|cccc|cccc}\toprule
                       &             & \multicolumn{4}{c|}{Random}                                                   & \multicolumn{4}{c}{PS}                   \\
  \midrule
                       &             & \#1              & \#2               & \#3               & \#4               & \#1       & \#2      & \#3     & \#4     \\
  \midrule
  \multirow{2}*{Airc.} & Accuracy    & 68.2             & 65.7              & 67.9              & 68.6              & 70.3      & 70.0     & 70.9    & 71.1    \\
                       & Active\_Pos & (1,7,12,22)        & (5,7,15,16)         & (5,7,8,19)          & (6,13,15,21)        & \multicolumn{4}{c}{(18,19,21,24)}          \\
  \midrule
  \multirow{2}*{DTD}   & Accuracy    & 54.8             & 54.6              & 55.9              & 56.7              & 57.0      & 56.0     & 56.9    & 56.7    \\
                       & Active\_Pos & (2,3,6,7,11,13,17) & (1,3,9,10,12,16,17) & (5,8,9,12,18,23,24) & (4,8,9,11,16,17,21) & \multicolumn{4}{c}{(18,19,20,21,22,23,24)} \\
  \bottomrule
  \end{tabular}
  }
\end{table*}

\begin{table}[]
  \centering
  \caption{Effects of $\lambda_s$ in Equation \ref{loss}. Active\_Pos Means the Positions of Activated Adapters.}
  \label{adaptive-weight}
  \setlength\tabcolsep{12pt}{
  \begin{tabular}{lcc}
    \toprule
    Settings                                                                 & Acitive\_Pos             & T2\_Acc (\%)   \\
    \midrule
    ${\lambda}_s$=$0.05$                                                             &/                  & 95.3       \\
    ${\lambda}_s$=$0.01$                                                              &1,2               & 95.4       \\
    ${\lambda}_s$=$0.001$                                                            &1-13               & 96.0       \\
    ${\bm \lambda}^\prime_s$=$adaptive$                                                   &11,12             & 96.3       \\
    \bottomrule
  \end{tabular}
  }
\end{table}

\begin{table}[]
  \centering
  \caption{Effects of $\lambda_a$ in Equation \ref{loss}. Active\_Pos Means the Positions of Activated Adapters.}
  \label{gate-weight}
  \setlength\tabcolsep{12pt}{
  \begin{tabular}{lcc}
    \toprule
    Settings                                                                 & Acitive\_Pos     & T2\_Acc (\%)   \\
    \midrule
    ${\lambda}_a$=$0.005$                                                    &11                 & 95.9      \\
    ${\lambda}_a$=$0.001$                                                    &11,12              & 96.3      \\
    ${\lambda}_a$=$0.0005$                                                   &11,12              & 96.0      \\
    ${\lambda}_a$=$0.0001$                                                   &11                 & 95.7       \\
    \bottomrule
  \end{tabular}
  }
\end{table}

\subsubsection{\bf Effects of Feature Adapter}
As feature adapter plays an essential role in acquiring new concepts, we expect it to be both powerful and lightweight.
Different types of feature extractor are verified for a good choice of the module, as shown in TABLE \ref{adapter-abla}.

\emph{plain}$_{1\times1}$ means the new features are linear combinations of existing features, while \emph{residual}$_{1\times1}$ means that the new features are the element-wise sum of existing features and adapted features.
It can be clearly seen that the proposed adapters boost a lot under the same parameter scale.

\subsubsection{\bf Visualization of Gates in Pruning Strategy}
To analyze the roles of \emph{gate-based pruning strategy}, we visualize the gate execution modes on VDD, as shown in Fig. \ref{gate_visualization}.
We observe that most tasks tend to activate gates in the deeper layers rather than shallower layers, like the \emph{conv3\_x} layers in ResNet.
This is because tasks in VDD prefer to share low-level features while adapt middle/high-level features in task-specific layers to obtain better performance.
We notice that the gate-activated number of CIFAR100 is much larger than the other tasks.
Because CIFAR100 belongs to the natural domain and contains a variety of images, which is more complicated than other tasks, needing more flexible structures for adapting the old model.
\begin{figure}[!hbtp]
\centering
\includegraphics[width=9cm, height=4cm] {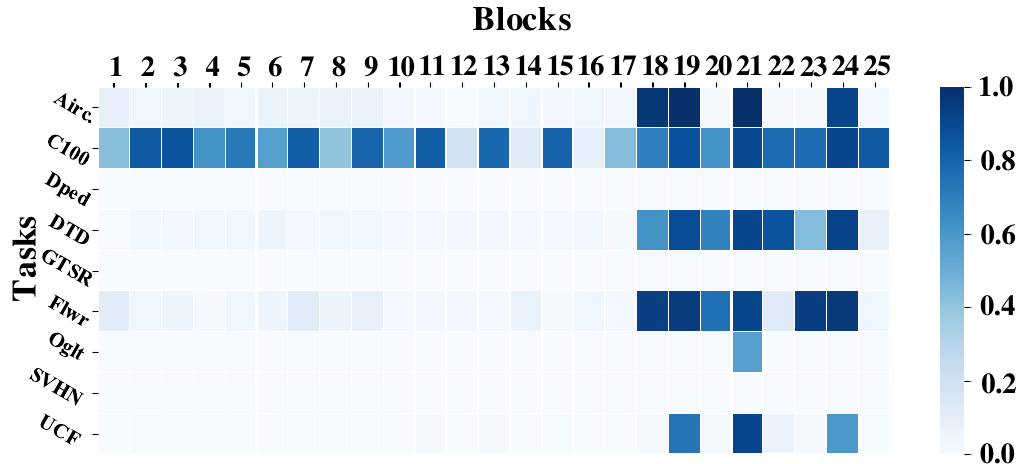}
\caption{The  gate activation in VDD. Color depth means activation probability.}
\label{gate_visualization}
\end{figure}

\subsubsection{\bf The Stability of Pruning Strategy}
To verify the stability of \emph{gate-based pruning strategy}, we conduct four rounds of experiments on VDD with different random seeds, as shown in TABLE \ref{active-gate-pos-analysis}.
Correspondingly, four control settings with randomly activated gates are verified.
We find our method is able to explore suitable activation positions to obtain better performance, \textit{i.e.}, (18,19,21,24) in Aric.
Notably, accuracy with random settings varies largely, while our method achieves more robust results.

\begin{figure*}
    \centering
    \includegraphics[width=16cm, height=3.5cm]{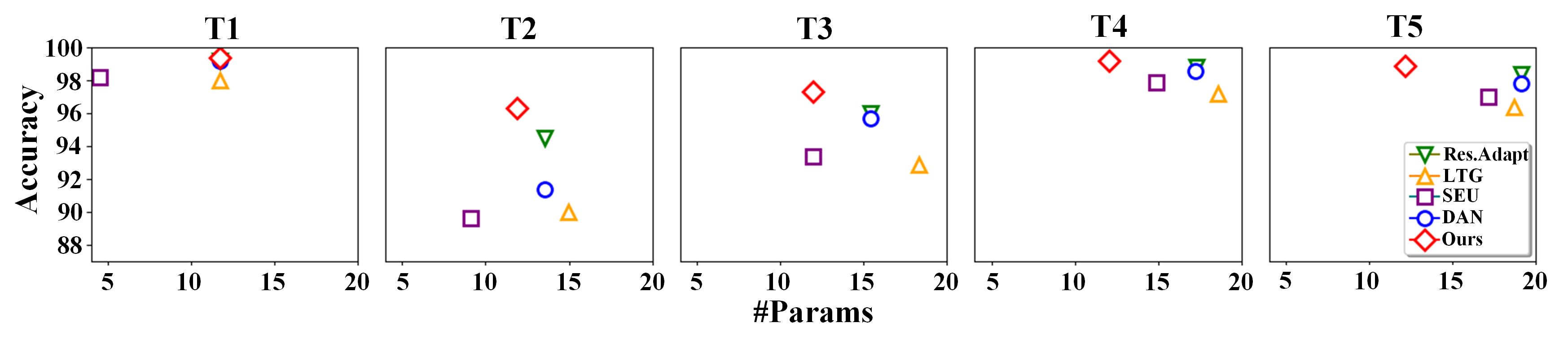}
    \caption{params-growth-curve. Parameters ($\times 10^6$) growth comparison of different methods. Methods lying in the top left corner refer to better results considering to the balance between performance and parameter scales.}
    \label{fig_first_case}
\end{figure*}

\begin{figure}
    \centering
    \includegraphics[width=5.5cm, height=4cm]{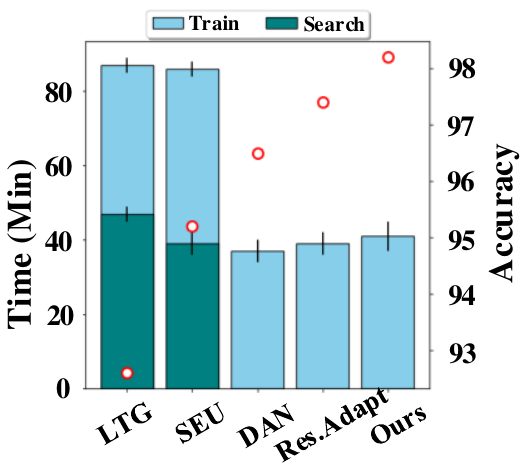}
    \caption{Time-consume comparison, including the Search and Train cost of different methods. Red circles means the accuracy.}
    \label{fig_second_case}
\end{figure}

\subsubsection{\bf Effects of the Adaptive $\lambda_s$ and $\lambda_a$}
We conduct experiments to show the effectiveness of adaptive training strategy on T2 of CIFAR-10, as shown in TABLE~\ref{adaptive-weight}.
For the constant ${\lambda}_s$, the model struggles to balance the accuracy and the active number of gates, resulting in suboptimal results.
In fact, different convolutional layers rely on different $\lambda_s$, \textit{i.e.}, the adaptive weights $\lambda_s^\prime$.
With $\lambda_s^\prime$, our model further improves the accuracy while keeping the minimal number of activated gates.
The hyper-parameter $\lambda_a$ is also analyzed in TABLE \ref{gate-weight}. Results show that our approach is not sensitive to $\lambda_a$ and therefore we choose 0.001 as the default.

\subsubsection{\bf Computational Cost} We here compare the computational cost of our methods with previous methods on Split CIFAR-10, as shown in Fig. \ref{fig_first_case}. 
For accuracy comparison with
SEU\cite{wang2020lifelong}, LTG\cite{li2019learn}, DAN\cite{rosenfeld2018incremental} and Res. Adapt\cite{2017LearningAdapters}, our method achieves the best results.
Compared to the parameter scale, our method only falls behind SEU in Task1 and Task 2, while the parameter scale of SEU quickly grows along with the increase of tasks (Task4 and Tasks 5). It is impractical to deploy methods like SEU on edge devices.
We also compare the cost performance of obtaining an incremental model with different methods.
Fig. \ref{fig_second_case} shows the average time as well as average accuracy on 5 tasks of Split CIFAR-10.
We find that NAS-based methods like SEU and LTG spend much search time compared to their training time,
while our method is very cost-efficient with high accuracy.

\begin{table*}[htp]
  \setlength\tabcolsep{2.7pt}
  \centering
  \caption{Accuracy Results on Split CIFAR-10. The Added Parameters ($\times 10^6$) of the Final Model and the Average Multiply-Add Counts (Mac) ($\times 10^6$) of 5 Tasks Are Also Provided. `+' Refers to the Added Parameters and Mac. `-' Means None. E$^2$-AEN \emph{(w/o P)} Means Our Method Without Pruning. 
  }
  \label{split-cifar10}
  \fontsize{7.5}{7.5}\selectfont
  \setlength\tabcolsep{14pt}{
  \begin{tabular}{lcccccccc}
    \toprule
    Method                                & T1    & T2    & T3    & T4    & T5    & AVG   & \#Params    & Mac \\
    \midrule
    joint\_train(Upper Bound)                     & 99.6  & 96.4  & 97.9  & 99.5  & 98.3  & 98.3  & 11.7      & 488.6    \\ 
    \midrule
    EWC-On \cite{2018Progress}           & 75.8  & 80.4  & 80.3  & 95.2  & 96.0    & 85.5  & $+$0      & $+$0    \\
    LwF \cite{2017Learning}              & 94.8  & 87.3  & 67.1  & 50.5  & 51.4  & 70.2  & $+$0      & $+$0    \\
    HAT \cite{J2018Overcoming}           & 98.8  & 91.1  & 95.3  & 98.5  & 97.7  & 96.3  & -         & $+$0 \\
    CCG \cite{abati2020conditional}           & 99.4  & 91.7  & 95.0    & 98.3  & 97.8  & 96.4  & -         & $+$0\\
    \midrule
    DAN$^1$ \cite{rosenfeld2018incremental}  & 99.2 & 91.4   & 95.7    & 98.6  & 97.8  & 96.5      & $+$7.4      & $+$ 69.8    \\ 
    Res. Adapt$^1$ \cite{2017Learning}       & 99.2  & 94.5  & 96.0    & 98.8  & 98.4  & 97.4   & $+$7.4      & $+$ 69.8    \\ 
    LTG$^1$ \cite{li2019learn} & 98.0    & 90.0    & 92.9  &97.2   &96.4   &94.9   &$+$7.0      &-   \\
    SEU$^2$ \cite{wang2020lifelong}           & 98.2  & 89.6  & 93.4  & 97.9  & 97.0    & 95.2  & $+$5.4      & -         \\
    E$^2$-AEN (w/o P)                   & 99.4  & 95.9  & 96.9  & {\bf 99.4}  & 98.7  & 98.1 & $+$7.4       & $+$ 69.8    \\ 
    E$^2$-AEN                    & {\bf 99.4} &{\bf 96.3}    &{\bf 97.3} & 99.2  & {\bf 98.9}   &{\bf 98.2} &{\bf $+$0.40} &{\bf $+$2.7}\\    
    \bottomrule
  \end{tabular}
  }
\end{table*}
\begin{table*}[t]
  \centering
  \caption{Top-1 Accuracy on the Visual Decathlon Challenge Test Set. \#P Is the Number of Parameters As a Factor
  of a Single-Task Implementation. `*' Means the Method Is Trained on All Datasets Jointly. }
  \label{results-on-vdd}
  \resizebox{\textwidth}{20mm}{
  \begin{tabular}{lclllllllllllc}
    \toprule
    method      & \#P  & ImNet & Airc. & C100  & Dped  & DTD   & GTSR  & Flwr  & Oglt  & SVHN  & UCF   & Mean   & Score \\
    \midrule
    Scratch \cite{2017LearningAdapters}   &10    & 59.87  &57.10  &75.73  &91.20  &37.77  &96.55  &56.3   &88.74  &96.63  &43.27  &70.32   &1625   \\
    Finetune\cite{2017LearningAdapters}  &10    & 59.87  &60.34  &82.12  &92.82  &55.53  &97.53  &81.41  &87.69  &96.55  &51.20  &76.51   &2500   \\
    LwF\cite{2017Learning}       &10    &59.87 &61.15 &82.23 &92.34 &58.83 &97.57 &83.05 &88.08 &96.1 &50.04 &76.93 &2515 \\
    \midrule
    Feature\cite{2017LearningAdapters}   &1     & 59.67  &23.31  &63.11  &80.33  &45.37  &68.16  &73.69  &58.79  &43.54  &26.8   &54.28   &544    \\
    BN adapt. \cite{2017Universal} &1.01    &59.87 &43.05 &78.62 &92.07 &51.60 &95.82 &74.14 &84.83 &94.10 &43.51 &71.76 &1363 \\
    \midrule
    Res. Adapt\cite{2017LearningAdapters} & 2    & 59.67  &56.68  & 81.2  & 93.88 & 50.85 & 97.05 & 66.24 & 89.62 & 96.13 & 47.45 & 73.88  & 2118  \\
    DAN \cite{rosenfeld2018incremental}       & 2.17 & 57.74  &64.12  & 80.07 & 91.3  & 56.54 & 98.46 & 86.05 & 89.67 & 96.77 & 49.38 & 77.01  & 2851  \\
    Piggyback \cite{2018Piggyback} &1.28  & 57.69  &65.29  &79.87  &96.99  &57.45  &97.27  &79.09  &87.63  &97.24  &47.48  &76.60   &2838   \\
    MTAN$^*$ \cite{2019End}     & 1.74 & 63.90  &61.81  & 81.59 & 91.63 & 56.44 &98.80  &81.04  &89.83  &96.88  &50.63  &77.25   &2941   \\
    {E$^2$-AEN}   & 1.39 & 62.08  &70.8$\pm$0.5  & 78.8$\pm$0.1 & 97.1$\pm$0.4 & 56.9$\pm$0.6 & 98.5$\pm$0.2 & 82.1$\pm$0.9 & 88.7$\pm$0.4  & 95.4$\pm$0.2 & 48.6$\pm$0.9 & {\bf 77.9$\pm$0.1}  & {\bf 3129$\pm$103}  \\

    \bottomrule
  \end{tabular}
  }
\end{table*}

\subsection{Comparisons on Classification Tasks}
Most methods only report their results on the classification benchmarks likc CIFAR-10 and VDD.
We first compare our methods with them.
\subsubsection{\bf Evaluation on Split CIFAR-10}
ResNet-18 is firstly trained on ImageNet as a backbone and is frozen when training all incremental tasks. 
We trained all models using 2 GPUs for 100 epochs with ADAM with an initial learning rate set to 0.002 and reduced every 40 epochs by a factor of 0.1. The weight decay is $1e^{-4}$.
The training strategy follows \cite{rosenfeld2018incremental, 2017LearningAdapters}.

TABLE~\ref{split-cifar10} shows the results of five incremental tasks on Split CIFAR-10. 
The parameter scale of the final task and the average multiply-add operations (MAC count) are also reported.
We find that E$^2$-AEN with \emph{feature adapters} is significantly superior to the regularization-based methods like EWC-On \cite{2018Progress} and LwF \cite{2017Learning}, and also achieves better results than the mask-based methods like HAT \cite{J2018Overcoming} and CCG \cite{abati2020conditional}.
Compared to the recent expandable networks like DAN \cite{rosenfeld2018incremental} and Res. Adapt \cite{2017LearningAdapters}, E$^2$-AEN without pruning strategy shows better performance, owing to the enhanced representation capacity of \emph{feature adapters}.
After equipping \emph{feature adapters} with \emph{gate-based pruning strategy}, E$^2$-AEN still achieves remarkable accuracy results (without any accuracy dropping).
E$^2$-AEN is lightweight and parameter-efficient, adding only $0.4\times 10^6$ extra parameters and $2.7\times 10^6$ Mac which is largely superior to existing dynamic expandable network SEU. 
Notably, our method achieves comparable results (98.2\%) with the upper bound \emph{joint\_train (98.3\%)}. 

\footnotetext[1]{DAN, Res. Adapt and LTG are re-implemented based on their original paper.}
\footnotetext[2]{The results of SEU are obtained based on their source code.}

\subsubsection{\bf Evaluation on Visual Decathlon Challenge}
We use the validation dataset to find the optimal training parameters, and then train the model on train and validation dataset for the final results.

We first train the model on ImageNet for 150 epochs, optimized by SGD with an initial learning rate of 0.1 and reduced every 35 epochs by a factor of 10.
For the sequential tasks, ADAM is used for 80 epochs with an initial learning rate of 0.01 and reduced by a factor of 0.1 in 40, 60 epochs respectively. The weight decay is $5e^{-4}$.
Note that, the data of other tasks is unavailable and parameters of previous tasks are frozen, and only AES and its prediction layers are learnable.
Details of the challenge can be found at the official website$^3$.
TABLE~\ref{results-on-vdd} shows the Top-1 classification accuracy and parameters compared to a single-task implementation.
As expected, our method achieves the new state-of-the-art.

\subsection{Comparisons on Object Detection Tasks}
Few works study IL problem on object detection tasks, which is not proportional to its wide applications in industry.
We here address the generalization of our method on object detection tasks.
We first evaluate our method on general object detection benchmarks (\emph{i.e.,} VOC and COCO), and then verify it on ICCV2021 SSLAD-Track3B challenge task.
\begin{table}[!htbp]
  \centering
  \caption{Detection Results (mAP) on COCO and VOC.}
  \label{detection}
  \setlength\tabcolsep{16pt}{
  \begin{tabular}{llllll}
    \toprule
    Methods        & COCO (\%)        & VOC (\%)  \\
    \midrule
    Joint Training & 47.8         & 74.1 \\ \midrule
    Fine-tuning    & 20.1 (-27.7) & 75.9 \\
    EWC            & 27.2 (-20.6)   & 75.0 \\
    AFD \cite{2019End}  & 27.8 (-20.0) & 77.7 \\
    E$^2$-AEN     & \textbf{47.8} (-0.0)  & \textbf{80.8} \\
    \bottomrule
  \end{tabular}
  }
\end{table}

\begin{table}[]
\centering
\caption{Evaluation on the ICCV-2021 SSLAD Challenge. The Models Are Evaluated on the Validation Set.}
\label{sslad_ret}
\setlength\tabcolsep{3pt}{
\begin{tabular}{cccc}
\toprule
Method & Detector    & Backbone     & Mean AP (\%) 
\\ \midrule
Rank1-COLT(ours)   & CascadeRCNN & Transformer     &75.11 
\\ \midrule

Rank2      & -  &-              & 62.27    
\\
Rank3      & - & -              & 59.56    
\\
Replay      & FasterRCNN & ResNet50               & 60.52 
\\
E$^2$-AEN      & FasterRCNN & ResNet50               &61.4 
\\ \midrule
\end{tabular}
}
\end{table}

\subsubsection{\bf General Object Detection}
Following \cite{2019End}, we consider the incremental task of COCO to VOC.
For COCO, we train the model on COCO valminusminival and tested on minival, 
in which the learning rate is 0.01 for 10 epochs and 0.001 for the last 2 epochs.
For VOC, we train the model on VOC2007 and VOC2012 train/val set, and test on VOC2007 test set, in which the initial learning rate is 0.002 and reduced by a factor of 0.1 after 10 epochs. 
Results in TABLE \ref{detection} show that our method achieves satisfying results on VOC without any harmless on COCO, demonstrating the effectiveness of our proposed approach across different architectures.


\subsubsection{\bf Evaluation on ICCV-2021 SSLAD Challenge}
The SSLAD-Track 3B challenge$^4$ at ICCV-2021 requires to design continual learning algorithms for object detection in automatic driving scenarios.
There are 4 different scenarios totally, including different times (daytime, night), different scenes (citystreet, highway) and different weathers (clear, overcast or rain).
In each scenario, there are six classes of objects, pedestrian, cyclist, car, truck, tram, and tricycle.
In the challenge, mean AP is used as the evaluation protocol.

We apply E$^2$-AEN to solve this incremental challenge, and evaluate it on the validation dataset. 
Results in Table \ref{sslad_ret} show that our method surpasses the rank3 method, and is also higher than the replay method. 
Note that, the reported results in Table \ref{sslad_ret} relied on the old data stored in buffer, while our method doesn't use them. 
By equipped with CascadeRCNN\&Transformer and the using of old data, the upgraded method (named as COLT \cite{sslad_report}) achieves the best results  and win first place. 
Details are shown in the technical report \cite{sslad_report}.

\footnotetext[3]{\url{https://www.robots.ox.ac.uk/~vgg/decathlon/}}
\footnotetext[4]{\url{https://competitions.codalab.org/competitions/33993\#learn_the_details}}

\subsection{Limitations}
Similar to most dynamic expandable works, our method also depends on a universal representation of the backbone which is pretrained on the large scale dataset, such as ImageNet.
In other words, if the universal representation of the backbone is not satisfied, more new parameters would be added to find a better performance.


\section{Conclusions}
 In this paper, we propose an end-to-end trainable adaptively expanding network named E$^2$-AEN.
This framework can dynamically generate lightweight network structures for new tasks without any accuracy drop in previous tasks.
Concretely, E$^2$-AEN has a serial of powerful \emph{feature adapters} and an adaptive \emph{gate-based pruning strategy}.
The former is used for preserving task-specific features coping with \emph{catastrophic forgetting}, while the latter is equipped with \emph{feature adapters} to adaptively generate lightweight structures, serving as a prune gate. The whole framework is efficient, and can be built upon any feed-forward architectures in a global optimization manner.
Extensive experiments on both classification and detection benchmarks demonstrate the effectiveness of our method. 
In the future, we will further explore the trainable expandable network.

\bibliographystyle{plain}
\bibliography{reference}

\vfill

\end{document}